%% file: main.tex
\documentclass[letterpaper, 10 pt, conference]{ieeeconf} 
                                                       
\IEEEoverridecommandlockouts                          
                                                        
\usepackage[english]{babel}
\usepackage[T1]{fontenc}

\usepackage[nolist]{acronym}
\usepackage[ruled,vlined]{algorithm2e}
\usepackage[cmex10]{amsmath}
\usepackage{amssymb}
\usepackage{bm}
\usepackage{booktabs}
\usepackage{color}
\usepackage[font=footnotesize,skip=1pt]{caption}
\usepackage{float}
\usepackage{graphicx,subfigure}
\usepackage{hyperref}
\usepackage{import}
\usepackage{mathtools}
\usepackage{multirow}
\usepackage{outline}
\usepackage{paralist}
\usepackage{siunitx}
\usepackage{soul}
\usepackage{stmaryrd}
\usepackage{systeme}
\usepackage{units}
\usepackage{url}
\usepackage{tikz}
\usepackage{booktabs}
\usetikzlibrary{tikzmark}

\input{macros}

\title{\LARGE \bf In-flight range optimization of multicopters using multivariable extremum seeking with adaptive step size}
\author{Xiangyu Wu and Mark W. Mueller
\thanks{Authors are with the High Performance Robotics Laboratory (HiPeRLab) at the Department of Mechanical Engineering, UC Berkeley. {\tt\small \{wuxiangyu,mwm\}@berkeley.edu}}
}

\begin{document}
\maketitle

\begin{abstract}
Limited flight range is a common problem for multicopters. 
To alleviate this problem, we propose a method for finding the optimal speed and heading of a multicopter when flying a given path to achieve the longest flight range. 
Based on a novel multivariable extremum seeking controller with adaptive step size, the method (a) does not require any power consumption model of the vehicle, (b) can adapt to unknown disturbances, (c) can be executed online, and (d) converges faster than the standard extremum seeking controller with constant step size.
We conducted indoor experiments to validate the effectiveness of this method under different payloads and initial conditions, and showed that it is able to converge more than $30 \%$ faster than the standard extremum seeking controller.
This method is especially useful for applications such as package delivery, where the size and weight of the payload differ for different deliveries and the power consumption of the vehicle is hard to model.

\end{abstract}

\section{Introduction}
\input{Sec1-Introduction.tex}
\label{sec:Introduction}

\section{Multicopter modeling}
\input{Sec2-Modeling.tex}

\section{Multivariable extremum seeking with adaptive step size}
\input{Sec3-Method.tex}

\section{Experimental evaluation}
\input{Sec4-Results.tex}

\section{Conclusion and future works}
\input{Sec5-Conclusion.tex}

\section*{Acknowledgements}
This work was supported by the Defense Advanced Research Projects Agency (DARPA) Subterranean Challenge. 
The experimental testbed at the HiPeRLab is the result of the contributions of many people, a full list of which can be found at \url{hiperlab.berkeley.edu/members/}.

{
\bibliographystyle{IEEEtran}
\bibliography{bibliography}
}

\begin{acronym}
\acro{HP}{high-pass}
\acro{LP}{low-pass}
\acro{ES}{Extremum Seeking}
\acro{MPC}{Model Predictive Controller}
\acro{UAV}{Unammmed Aerial Vehicle}
\acro{MAV}{Micro Aerial Vehicle}
\acro{TSMP}{Traveling Sales Man Problem}
\acro{VTOL}{Vertical Take Off and Landing}
\acro{ASL}{Autonomous Systems Lab.}
\acro{IDSC}{Institute for Dynamic Systems and Control}
\acro{ICRA}{International Conference in Robotics and Automation}
\acro{COM}{Center of Mass}
\acro{ESC}{Electronic Speed Controller}
\acro{MoCap}{Motion-Capture}
\acro{OCP}{Optimal Control Problem}
\acro{OC}{Optimal Control}
\acro{EKF}{Extended Kalman Filter}
\acro{CFD}{Computational Fluid Dynamics}
\acro{CV}{constant velocity}
\end{acronym}

\end{document}

%% file: macros.tex
\newcommand{\mnorm}[1]{\left\| #1 \right\|}

\newcommand{\mvec}[1]{\boldsymbol{#1}}
\newcommand{\mvecd}[1]{\dot{\boldsymbol{#1}}}
\newcommand{\mvecdd}[1]{\ddot{\boldsymbol{#1}}}

\newcommand{\mmat}[1]{\begin{bmatrix} #1 \end{bmatrix} }

\newcommand{\mrb}[1]{\left( #1 \right)} 

\newcommand{\figref}[1]{Fig.~\ref{#1}}

\newcommand{\crossMat}[1]{\ensuremath{S(#1)}}

\newcommand{\skewSym}[1]{\ensuremath{\,S\!\mrb{#1}}}

%% file: Sec1-Introduction.tex
Multicopters are used in a wide range of applications such as aerial photography \cite{AerialPhotograpy}, transportation \cite{tagliabue2017collaborative}, search and rescue \cite{bahnemann2017decentralized} and inspection \cite{Inspection}, thanks to their simplicity in fabrication and control, high maneuverability and low cost.
However, the limited flight range of most available platforms \cite{karydis2017energetics} severely constrains their range of applications.

\par 
One way to address the limited flight autonomy problem is novel hardware design.
For example, \cite{TriangularQuad} designed a triangular quadrotor, which has one large rotor for lift and three small rotors for control, combining the energy efficiency of the large rotor and the fast control response of the small rotors; 
\cite{kalantari2013design} designed a hybrid quadcopter which is able to do both aerial and ground locomotion. When the vehicle operates in the ground locomotion mode, it only needs to overcome the rolling resistance and uses much less power compared to flying; \cite{switchingBattery} proposed a quadcopter capable of in-flight battery switching.

\par 
Algorithm-based optimization is another way to reduce multicopters' power consumption. 
By planning energy-efficient trajectories or by implementing energy-aware control algorithms, this approach does not require changes on existing hardware and is thus economic to deploy.
Algorithmic improvements can be achieved through model-based or model-free methods.
Model-based methods (e.g. \cite{modelBased1} and \cite{modelBased2}) benefit from fully exploiting the capabilities of the system, but are dependent on the availability of an accurate power consumption model of the system. 
The vehicle's power consumption model is usually derived from analyzing its electric and aerodynamic properties. For example, \cite{ampatis2014parametric} and \cite{uavEnergyModeling} introduced power consumption models of the battery, electric speed controller and motor, and \cite{leishman2006principles} introduced the aerodynamic power consumption of the propeller based on momentum theory.
Model-free methods (e.g. \cite{kreciglowa2017energy}), on the other hand, could take into account hard-to-model effects, such as changes in vehicle components' performance due to aging and changes in the aerodynamic drag of the vehicle due to carrying different payloads or different wind speed. 
Model-free methods are especially useful for applications such as package delivery, where the size and the weight of the package differ for different deliveries.
\begin{figure}[t]
	\centering
	\subfigure{\includegraphics[width=0.45\linewidth]{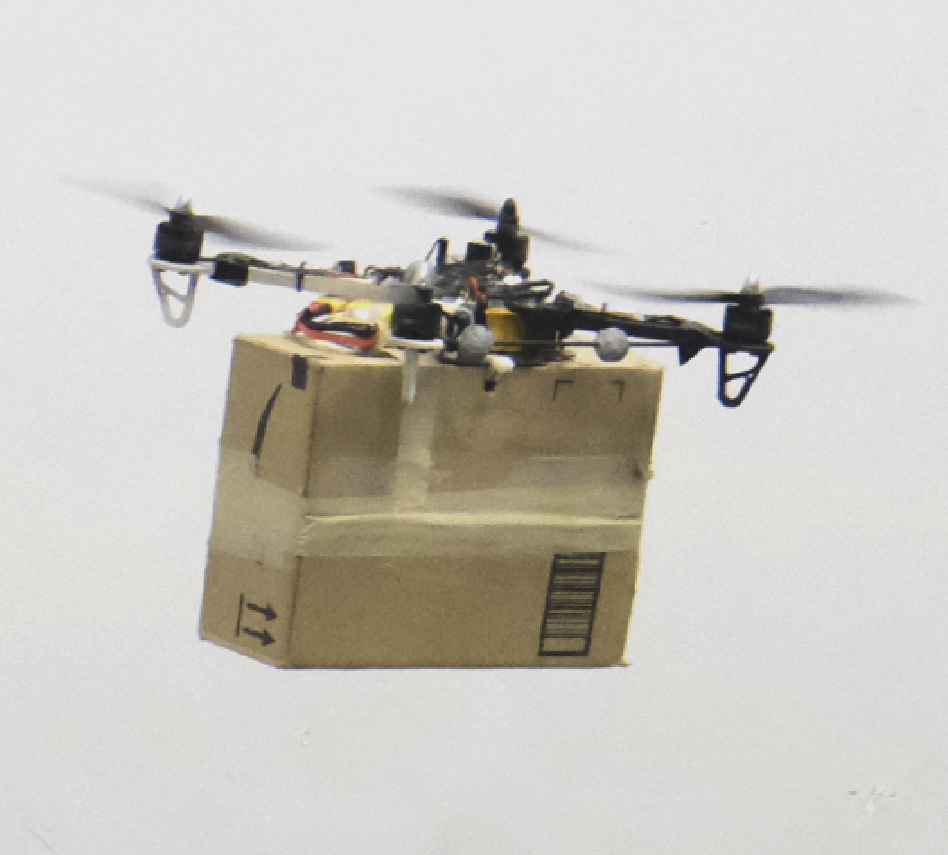}}
	\subfigure{\includegraphics[width=0.45\linewidth]{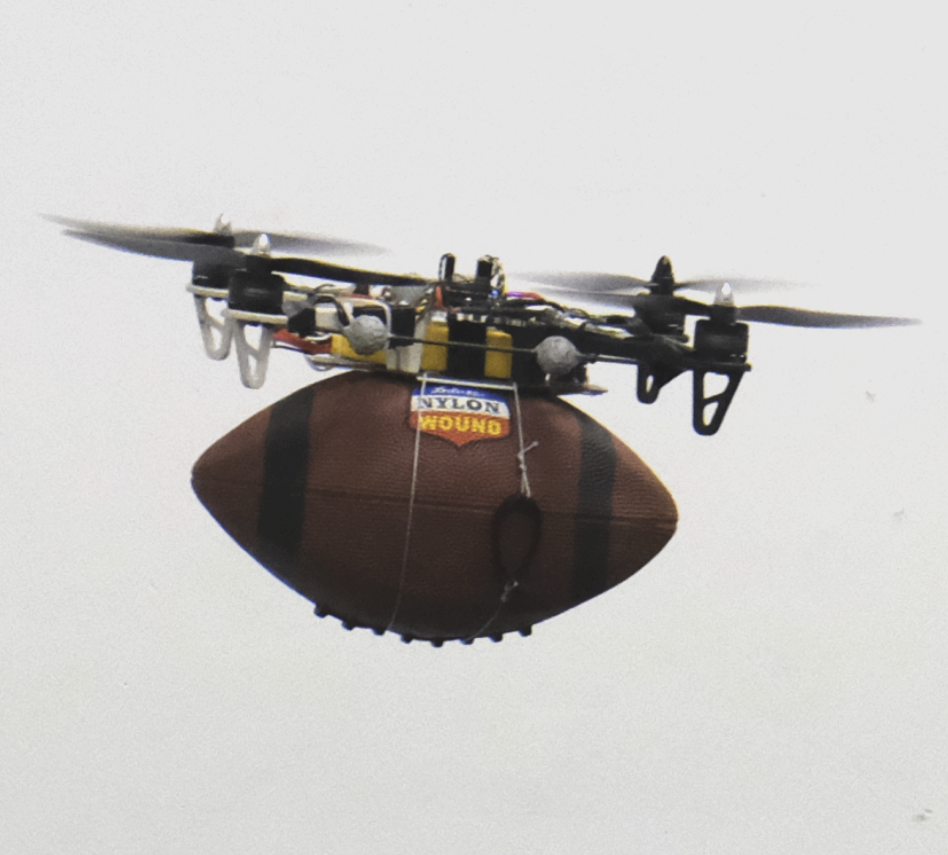}}
	\caption{ A flying quadcopter carrying different payloads. 
	The proposed extremum seeking controller with adaptive step size can find the speed and heading that maximize the flight range of the vehicle when flying along a user-defined geometric path. 
	Besides, it adapts to unknown disturbances, such as different payloads. Experiments were conducted to evaluate the performance of the proposed method, which can be seen in the video attachment.}
	\label{fig:front}
\end{figure}

\par
In this work, we present a model-free, online approach to find the speed and heading of a multicopter that maximizes its total flight distance (range) using a multivariable extremum seeking controller with adaptive step size. 
Path planning for multicopters typically has some additional (redundant) degrees of freedom (e.g. the speed and heading of the vehicle).
The proposed method autonomously sets the reference vehicle speed and heading along a predefined geometric path (e.g. a path for delivering a package from the warehouse to its destination) and parameterizes it into a reference trajectory, which is tracked by the low-level controller of the vehicle. 
The extremum seeking control is a technique for finding the input to a system that achieves the optimal output, and a detailed description and stability analysis can be found at \cite{multiESC1} and \cite{multiESC2}.
For example, to find the minimum-power flight speed of the vehicle, the extremum seeking controller applies a periodic perturbation to the reference speed. And by monitoring the corresponding changes in the vehicle's power consumption, it estimates the gradient of the vehicle's power with respect to speed and performs gradient descent to find the optimal speed. 

\par
Besides, unlike the standard extremum seeking controller, which does gradient descent with a constant step size, we propose a novel extremum seeking controller that adapts the step size based on the first and second moments of the gradient estimation. 
The idea comes from Adam \cite{Adam}, a popular stochastic optimization method for training neural networks.
Compared to the standard multivariable extremum seeking controller, the proposed adaptive step size method can achieve faster convergence.
In indoor experiments we show the effectiveness of the proposed method in finding the optimal speed and heading by comparing the results from the extremum seeking controller with ground truth data; we also show the performance improvements of this new method compared to the standard extremum seeking controller.

In summary, the main contributions of this work are: 
\begin{inparaenum}[(a)]
	\item the proposal of a novel extremum seeking controller with adaptive step size, which converges faster than the standard extremum seeking controller, by blending the ideas from the machine learning field,
	\item an extension of our previous work \cite{previousWork} of finding the optimal range speed of multicopters by taking into account the optimization of the vehicle's heading, and
	\item experimental validation of the proposed method and comparison with the standard extremum seeking controller.
\end{inparaenum}


%% file: Sec2-Modeling.tex
\label{sec:modeling}
In this section we define the reference frames, briefly introduce the dynamics and a power consumption model of the quadcopter.
The model is given to help to understand the effect of the vehicle's speed and sideslip on its power consumption. It is not used for the flight range optimization.

\subsection{Reference frame definition}
\begin{figure}[b]
	\begin{center}
		\includegraphics[width = 0.8\linewidth]{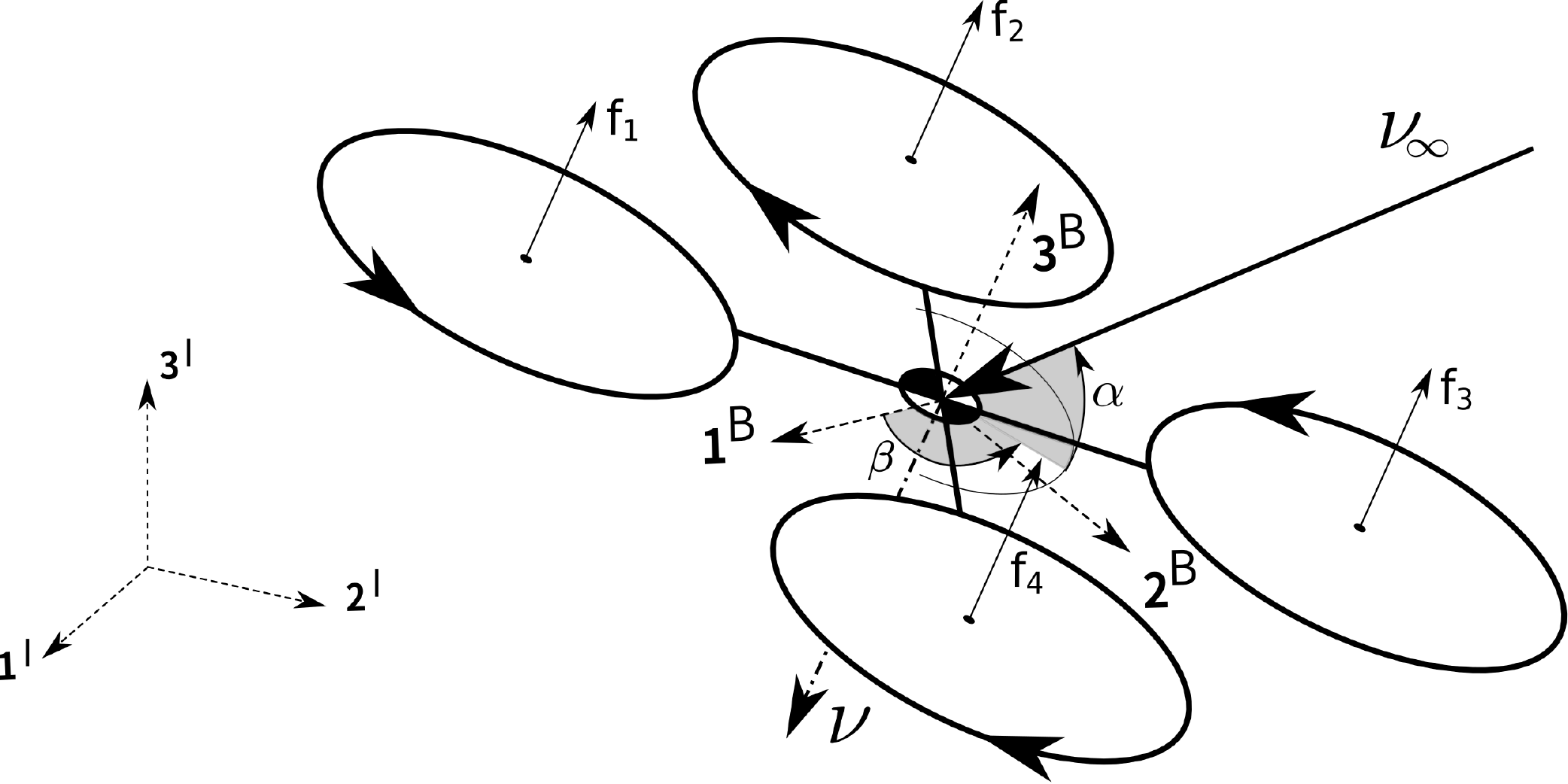}
	\end{center}
	\caption{
		$I$ represents the inertial reference frame and $B$ the quadcopter reference frame. We additionally show the thrust force of the $i$-th propeller $\mvec{f}_i$, the free-stream velocity $\mvec{\nu}_\infty$, the induced velocity $\mvec{\nu}$, angle of attack $\alpha$, and the sideslip $\beta$, shown positive in the diagram.
	}
	\label{figVehicleDiagram}
\end{figure}

As shown in  \figref{figVehicleDiagram}, two sets of reference frames are defined --  an inertial frame $I$ attached to the ground and a body-fixed frame $B$ attached to the Center of Mass (COM) of the quadcopter.

\subsection{Quadcopter dynamics}
The quadcopter is modeled as a rigid body with six degrees of freedom: three degrees of freedom from the linear translation along the three axes of the inertial frame $I$ and three degrees of freedom from the three-axis rotation from the body frame $B$ to the
inertial frame $I$, described by an orthogonal rotation matrix $\mvec{R}$.
Each propeller $i$ ($i = 1, ..., 4$), produces a thrust $\mvec{f}_i = (0,0,f_i)$ and a torque $\mvec{\tau}_i = (0,0,{\tau}_i)$, both expressed in $B$.

\par
According to  \cite{schulz2015high}, the drag force $\mvec{f}_d$ (expressed in $I$) is consisted of a linear term and a quadratic term
\begin{align}
\begin{split}
\mvec{f}_d := -\mrb{\mu_1 \nu_\infty + \mu_2 \nu_\infty ^2}\frac{\mvec{\nu}_\infty}{\mnorm{\mvec{\nu}_\infty}_2}
\label{eq:dragForce}
\end{split}
\end{align}
where $\mvec{\nu}_\infty$ corresponds to the free-stream velocity expressed in $I$, and $\mu_1$ and $\mu_2$ represent the drag coefficients, which are dependent on the vehicle sideslip, among others.
In addition, we assume that the drag force acts on the center of mass of the multicopter and thus no torque is produced because of it.

\par
The orientation of the vehicle with respect to the free-stream velocity $\mvec{\nu}_\infty$ is described by the angle of attack $\alpha$ and the sideslip $\beta$, as shown in \figref{figVehicleDiagram}. The angle of attack $\alpha$ is defined as the angle between ${\mvec{\nu}_\infty}$ and the plane given by $\mvec{1}^B$ and $\mvec{2}^B$; the sideslip is defined as the angle between  $\mvec{1}^B$ and the projection of ${\mvec{\nu}_\infty}$ in the plane given by $\mvec{1}^B$ and $\mvec{2}^B$. 

\par
With $\mvec{x}$ and its derivatives denoting the translational position, velocity and acceleration of the vehicle, $\mvec{g}$ denoting the gravity acceleration, all in the inertial frame $I$, the translational dynamics of the vehicle is given by
\begin{align}
m \mvecdd{x} &= m \mvec{g} + \mvec{R} \sum{\mvec{f}_i} + \mvec{f}_d \label{eq:DynamicsTranslation}
\end{align}
In addition, with $\mvec{\omega}$ and its derivative denoting the angular velocity and acceleration, expressed in $B$, the rotational dynamics of the vehicle is given by
\begin{align}
\mvecd{R} =& \mvec{R}\crossMat{\mvec{\omega}} \\
\mvec{I} \mvecd{\omega} =& -\mvec{\omega}\times\mvec{I}\mvec{\omega} + \sum \mvec{\tau}_i 
\end{align}
where $\mvec{I}$ is the mass moment of inertia and $\skewSym{\mvec{\omega}}$ is the skew-symmetric matrix form of the vector cross product such that 
\begin{align}
\skewSym{\mvec{\omega}} = \mmat{0 & -\omega_3 & \omega_2 \\ \omega_3 & 0 & -\omega_1\\-\omega_2 & \omega_1 &0}
\end{align}

\subsection{Power consumption}
Following \cite{karydis2017energetics}, \cite{ware2016analysis} and \cite{kreciglowa2017energy}, we assume that the total power consumption $P$ (measured at the terminals of the battery) is proportional to the aerodynamic induced power ${P}_\text{induced}$ \cite{leishman2006principles}
\begin{align}
{P} = & \frac{1}{\eta}{P}_\text{induced}
\label{eq:lumpedPowerConsumption}
\end{align}
where $\eta$ lumps together the efficiency of the motors and propellers.
In forward flight $P_\text{induced}$ is computed as 
\begin{align}
p_{\text{induced}} = \kappa\mrb{\nu + \nu_\infty \sin\alpha} \sum{f_i}
\label{eq:inducedPowerConsumption}
\end{align}
where $\nu_\infty$ = ${\mnorm{\mvec{\nu}_\infty}_2}$ is the magnitude of the free-stream velocity and $\nu$ represents the induced velocity applied by the propeller to the surrounding air.
The scalar $\kappa$ is an empirical correction factor.
The induced velocity $\nu$ is implicitly defined by
\begin{align}
\nu = \frac{\nu_h^2}{\sqrt{(\nu_\infty \cos \alpha )^2 + (\nu_\infty \sin \alpha + \nu)^2}}  \quad
\nu_h = \sqrt{\frac{mg}{8 \rho \pi r^2}}
\label{eq:InducedVelocity}
\end{align}
where $\rho$ is the density of the air and $r$ is the radius of the propellers. The induced velocity can be solved for numerically using techniques such as the Newton-Raphson method. A more detailed derivation and explanation of the power consumption model can be found in our previous work \cite{previousWork}.
When the vehicle is flying at a low speed, increasing the flight speed can help to increase the free-stream velocity $\nu_\infty$ and to decrease the induced power consumption. 
This effect was also observed by \cite{karydis2017energetics} and \cite{leishman2006principles}.
When the speed is further increased, the drag force $\mvec{f}_d$ becomes significant and large motor thrusts are needed for the flight, which causes power consumption to increase.
The sideslip also affects the vehicle's power consumption by affecting the drag force $\mvec{f}_d$ if the vehicle is not axisymmetric.
Thus, by affecting the power consumption of the vehicle, the speed and sideslip affects the vehicle's range.


%% file: Sec3-Method.tex
\label{sec:OnlineAdaptiveController}
In this section we introduce the multivariable extremum seeking controller which is able to find the optimal range speed and sideslip of a multicopter, given a user-defined geometric reference path.
This method does not depend on any power consumption model of the vehicle and can adapt to unknown disturbances such as different payloads.
Changes in the vehicle's speed and sideslip affect the power consumption and thus the range of the vehicle, which is experimentally demonstrated in Section \ref{sec:evaluation}.

\subsection{Cost function derivation}
\label{sec:costFunctionDefinition}
In this section we derive the cost function used in flight range optimization.
The cost function's derivation follows our previous paper \cite{previousWork}.
We assume that the multicopter 
\begin{inparaenum}[(a)]
	\item has energy $E \in [E_\text{empty}, E_\text{full}]$ during the operation, such as energy stored in the battery and does not have in-flight battery charging,
	\item is moving with constant speed $v := \left\lVert{\mvec{v}}\right\rVert_2$ and sideslip $\beta$, and
	\item is using constant electric power $P$.
\end{inparaenum}

The total flight distance $d_\text{flight}$ is defined as
\begin{align}
\begin{split}
d_\text{flight} & := \int_{t_0}^{t_\text{end}}v\ dt = \int_{E_\text{empty}}^{E_\text{full}} \frac{v}{P} dE  = \frac{v}{P} \Delta E
\label{eq:RangeCostFunction}
\end{split}
\end{align}
Thus, maximization of the flight range corresponds to minimization of  ${P}/{v}$
\begin{align} 
\label{eq:RangeCostFunctionFinal}
\max(d_\text{flight}) \Leftrightarrow \max \Bigg(\frac{v}{P} \Bigg) \Leftrightarrow  \min \Bigg(\frac{P}{v} \Bigg).
\end{align}
and ${P}/{v}$ is selected as the cost function for the optimal range extremum seeking problem.

\subsection{Adaptive step size multivariable extremum seeking controller}
\label{sec:esc}
\begin{figure}
	\begin{center}    
		\includegraphics[width=\linewidth]{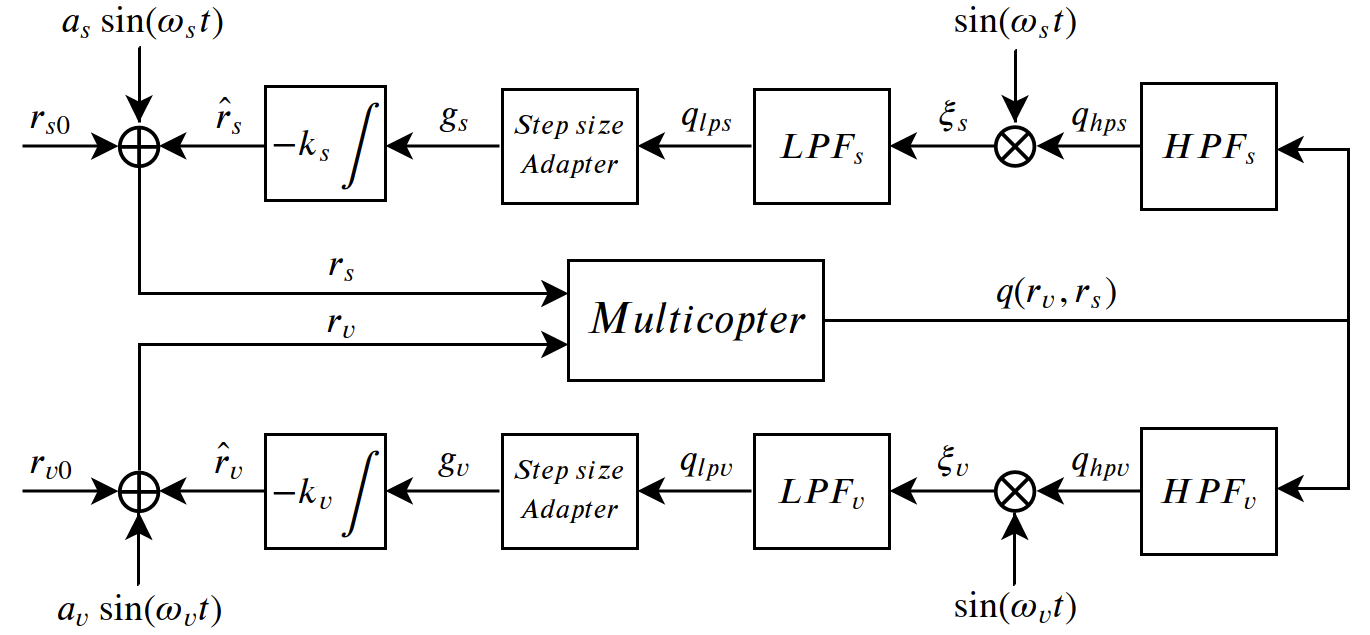}
	\end{center}
	\caption{The block diagram of the adaptive step size multivariable extremum seeking controller. The goal of the controller is to find the optimal $r_s$ and $r_v$ to minimize the cost function $q(r_v, r_s)$. The scalar $r_{v0}$ and $r_{s0}$ represent the plant's initial setpoints for speed and sideslip. 
	The frequency of the high-pass and low-pass filters is set, respectively, to $\omega_{\text{hpv}}$ and $\omega_{\text{lpv}}$ for speed and $\omega_{\text{hps}}$ and $\omega_{\text{lps}}$ for sideslip. 
	The scalar $k_v$ and $k_s$ are related to the step size of the extremum seeking controller and both of them should be positive numbers to minimize the cost function $q(r_v, r_s)$. The diagram of the standard extremum seeking controller does not have the step size adapter and the output of the low-pass filter directly goes to the integrator, while other parts of the diagram are exactly the same.
	}
	\label{fig:escDiagram}
\end{figure}

The proposed adaptive step size multivariable extremum seeking controller is shown in \figref{fig:escDiagram}.
It can find an unknown vehicle flight speed $r_v^*$ and sideslip $r_s^*$ that minimize the cost function $q(r_v, r_s)$ without depending on any power consumption model.
For the optimal range goal, the cost function is instantaneous power over speed.
The derivation of the cost function can be found in Section \ref{sec:costFunctionDefinition}.

\par
The extremum seeking controller approximates the gradient of the cost function and performs gradient descent to minimize the cost \cite{multiESC2}. To get the cost function's gradient with respect to speed, the extremum seeking control uses a dithering signal $a_v\sin(w_vt)$, a demodulation signal $\sin(w_vt)$, a high-pass filter $HPF_v$ and a low-pass filter $LPF_v$ to estimate the gradient of the cost function $q$ at the current set point $\frac{\partial q}{\partial v}(r_{v0}+\hat{r}_v)$.
We assume that the set point $r_{v0}+\hat{r}_v$ changes much slower than that of the dithering and demodulation signal.
The Taylor series expansion of the cost function is
\begin{align}
&q(r_{v0}+\hat{r}_v+a_v\sin(w_vt)) \nonumber \\ 
& =  q(r_{v0}+\hat{r}_v)+a_v\sin(w_vt)\frac{\partial q}{\partial v}(r_{v0}+\hat{r}_v) + O(r_{v0}+\hat{r}_v)
\end{align} 
where $O(r_{v0}+\hat{r}_v)$ stands for higher order terms.
The high-pass filter $HPF_v$ would suppress the low frequency signal $q(r_{v0}+\hat{r}_v)$, and thus the output of the high-pass filter, $q_{hpv}$, is
\begin{align}
a_v\sin(w_vt)\frac{\partial q}{\partial v}(r_{v0}+\hat{r}_v) + O(r_{v0}+\hat{r}_v)
\end{align}
Multiplication with the demodulation signal results in
\begin{align}
a_v\sin^2(w_vt)\frac{\partial q}{\partial v}(r_{v0}+\hat{r}_v) + \sin(w_vt)O(r_{v0}+\hat{r}_v)
\end{align}
which is defined as $\xi_v$. It has a low frequency component (or average) given by 
\begin{align}
R\frac{\partial q}{\partial v}(r_{v0}+\hat{r}_v)
\end{align}
where $R = \lim_{T\to\infty} \frac{1}{T}\int_{0}^{\infty} a_v\sin^2(w_vt) dt$. Thus, the output of the low-pass filter, $q_{lpv}$, is an estimate of the cost function's gradient with respect to speed.
For the same reason, the output of the low-pass filter $LPF_s$, $q_{lps}$, is an estimate of the cost function's gradient with respect to sideslip.

\par
The difference between the proposed extremum seeking controller and the standard multivariable extremum seeking controller is the step size adapter, whose pseudo-codes are shown in Algorithm \ref{alg:adapter}.
The idea comes from the adaptive moment estimation algorithm (Adam) \cite{Adam}, which is commonly used in the stochastic optimization of objective functions in machine learning, such as training neural networks. 

\begin{algorithm}
	\caption{Step size adapter used in the proposed multivariable extremum seeking controller. $\beta_1$ = 0.9, $\beta_2=0.999$, and $\epsilon$ = $10^{-8}$, which are the default Adam parameters.} \label{alg:adapter}
	{Input}{$\: q_{lp}$} ($q_{lpv}$ or $q_{lps}$), {Output}{$\: g$} ($g_v$ or $g_s$) \\
	Define and initialize $\beta_1$, $\beta_2$, $\epsilon$ and $threshold$ \\
	$m_0 \gets q_{lp0}$; (Initialize the first moment) \\
	$v_0 \gets q^2_{lp0}$; (Initialize the second moment)  \\
	$i \gets 0$; (Initialize timestep) \\
	\While {flight not end}{
		$i \gets i + 1$\;
		$m_i \gets \beta_1 * m_{i-1} + (1-\beta_1) * q_{lp}$\; 
		$v_i \gets \beta_2 * v_{i-1} + (1-\beta_2) * q^2_{lp}$\;
		\eIf{$\sqrt{v_i} > threshold$}{
			$g = m_i / (\sqrt{v_i}+\epsilon)$\;
		}{
			$g = m_i * (\sqrt{v_i}+\epsilon) / threshold^2$\;
		}
	}
\end{algorithm}

\par
The adapter takes in the output of the low-pass filter ($q_{lpv}$ for speed and $q_{lps}$ for sideslip), which is the gradient estimation and outputs $g$, which is passed to the integrator to perform gradient descent.
The effective step size for gradient descent is ${k_v}{g_v}/{q_{lpv}}$ for the speed optimization and ${k_s}{g_s}/{q_{lps}}$ for the sideslip optimization, and the step size adapter changes them by setting $g_v$ and $g_s$.
By taking the exponential moving average of the low-pass filter output, it helps to reduce the noise in the gradient estimation. 
The first output from the low-pass filter $q_{lp0}$ is used to initialize the moving average.
By dividing the first moment with the square root of the second moment ($\sqrt{v_i}$), the output $g$ of the adapter will be similar for the sideslip and the speed, leading to similar convergence speed for these two variables if $k_v$ and $k_s$ are set the same; in addition, it outputs a small value when the gradient estimate has a large uncertainty ($m_i$ is small) and vice versa, which is good for the stability of the extremum seeking controller. 
When the square root of the second moment $\sqrt{v_i}$ is less than the threshold defined, which means the extremum seeking controller is close to convergence, the output $g$ will decrease as $\sqrt{v_i}$ decreases to avoid oscillations of $\hat{r}_v$ and $\hat{r}_s$ near convergence.

\par
For the multivariable extremum seeking controller to work for both the speed and the sideslip simultaneously, the frequency of their disturbances $\omega_v$ and $\omega_s$ should be distinct \cite{multiESC2}. One choice of the parameters is detailed in the experimental evaluation section (Section \ref{sec:evaluation}).  

\begin{figure}
	\begin{center}    \includegraphics[width =  \linewidth]{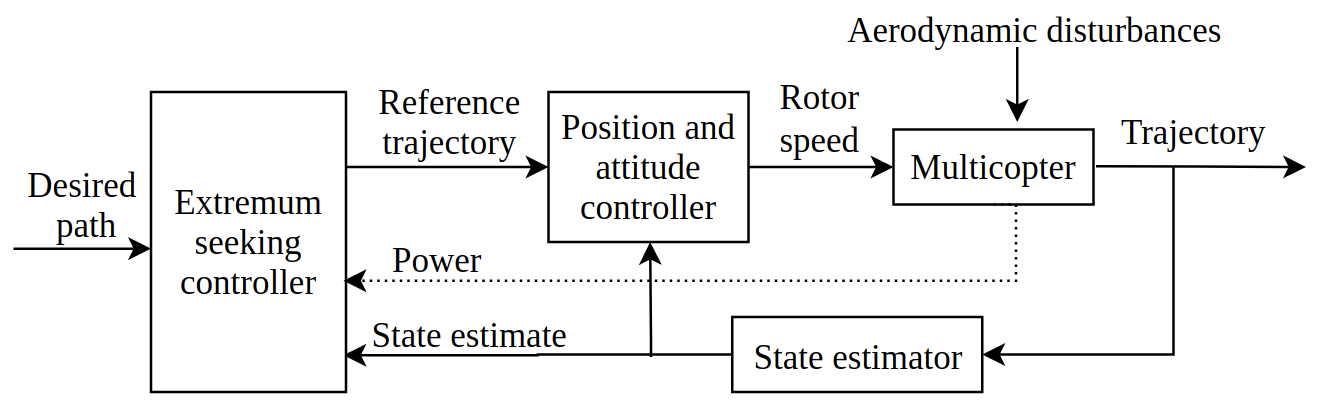}
	\end{center}
	\caption{System diagram for the model-free adaptive range optimization of a multicopter.}
	\label{fig:escFullDiagram}
\end{figure}

\par
The role of the extremum seeking controller in the vehicle control architecture is shown in \figref{fig:escFullDiagram}.
Given a desired geometric path, instantaneous power measurements and instantaneous speed and sideslip measurements, the extremum seeking controller sets the reference tangential speed and sideslip along the desired path and parameterizes it into the reference trajectory, which is tracked by the low-level position and attitude controller. The power measurements come from an onboard voltage and current sensor, and the speed and sideslip measurements come from a state estimator based on a motion capture system. 

%% file: Sec4-Results.tex
\label{sec:evaluation}
Indoor experiments were conducted to evaluate the performance of the proposed adaptive step size, multivariable extremum seeking method, and can be seen in the video attachment.
We show that the proposed method is able to find the optimal sideslip and speed for different payloads and can converge more than $30 \%$ faster than the standard extremum seeking controller. 

\subsection{Experimental setup}
A custom-built quadcopter, as shown in \figref{fig:front} was used during the experiments.
The vehicle weighs 660 grams without payload. 
The distance between the hubs of two diagonal motors is 330 mm and the propeller is 203 mm in diameter. 
We used two payloads during the experiment, one is a 255$\times$180$\times$85 mm cardboard box weighing 120 grams; the other one is an American football weighing 329 grams.
A Crazyflie 2.0 running a custom version of PX4 firmware was used as the low-level flight controller for the vehicle.  
The experiments were conducted in an indoor flight space with a size of 7$\times$6$\times$5 m. 
A motion capture system was used for the state estimation of the vehicle and a voltage and current measurement module was connected to the battery to measure the instantaneous power consumption of the vehicle.

\par
The value of parameters of the standard extremum seeking controller and the proposed adaptive step size extremum seeking controller used throughout the experiments are shown in Table \ref{tab:paramter}.
The perturbation frequency of reference speed $w_v$ was set to 1 rad/s, and 
the perturbation frequency of reference sideslip $w_s$ was set to 0.5 rad/s.
Multivariable extremum seeking control requires $w_v \neq w_s$ \cite{multiESC2}.
While increasing the perturbation frequencies is helpful to improve the convergence rate of the extremum seeking controller, one should make sure they are not too large for the vehicle to track. The cutoff frequencies of the high-pass and low-pass filters were set to be the same as their corresponding perturbation frequency. The magnitude of speed perturbation $a_v$ was set to be 0.15 m/s and the magnitude of sideslip was set to 7.5$^{\circ}$.
These values need to be selected large enough to provide the extremum seeking controller with gradient information of the cost function and also make the neighborhood around the optimal value that the extremum seeking controller converges to small. The $threshold$ parameter in Algorithm \ref{alg:adapter} was empirically set to be 1.

\par
To make a fair comparison, we kept all the control parameters for the two different methods to be the same except $k_v$ and $k_s$, since they have different meanings for the two methods: the $k_v$ and $k_s$ values are the step sizes for the standard method but are only part of the step sizes for the adaptive method, as shown in Section \ref{sec:esc}. 
They were empirically tuned in experiments for the two different methods to achieve the fastest convergence rate while guaranteeing the stability of the system (too large $k_v$ and $k_s$ will make the system unstable). 

\begin{table}
	\renewcommand{\arraystretch}{1.2}
	\caption{Values of extremum seeking parameters}
	\label{tab:paramter}
	\centering
	\begin{tabular}{|l|l|l|}
		\hline
		Parameter  		  & Proposed method & Standard method \\ \hline
		$a_v$             & 0.15 m/s  &  0.15 m/s  \\ \hline
		$w_v$             & 1 rad/s   &  1 rad/s   \\ \hline
		$w_{hpv}, w_{lpv}$         & 1 rad/s   &  1 rad/s   \\ \hline
		$k_v$			  & 0.1		   &  0.025		 \\ \hline

		$a_s$             & 7.5$^{\circ}$    &  7.5$^{\circ}$    \\ \hline
		$w_s$             & 0.5 rad/s   &  0.5 rad/s   \\ \hline
		$w_{hps}, w_{lps}$         & 0.5 rad/s   &  0.5 rad/s   \\ \hline
		$k_s$			  & 0.1		   &  0.02		 \\ \hline	
	\end{tabular}
\end{table}

\subsection{Performance evaluation}
\begin{figure}
	\centering
	\includegraphics[width=0.95\linewidth]{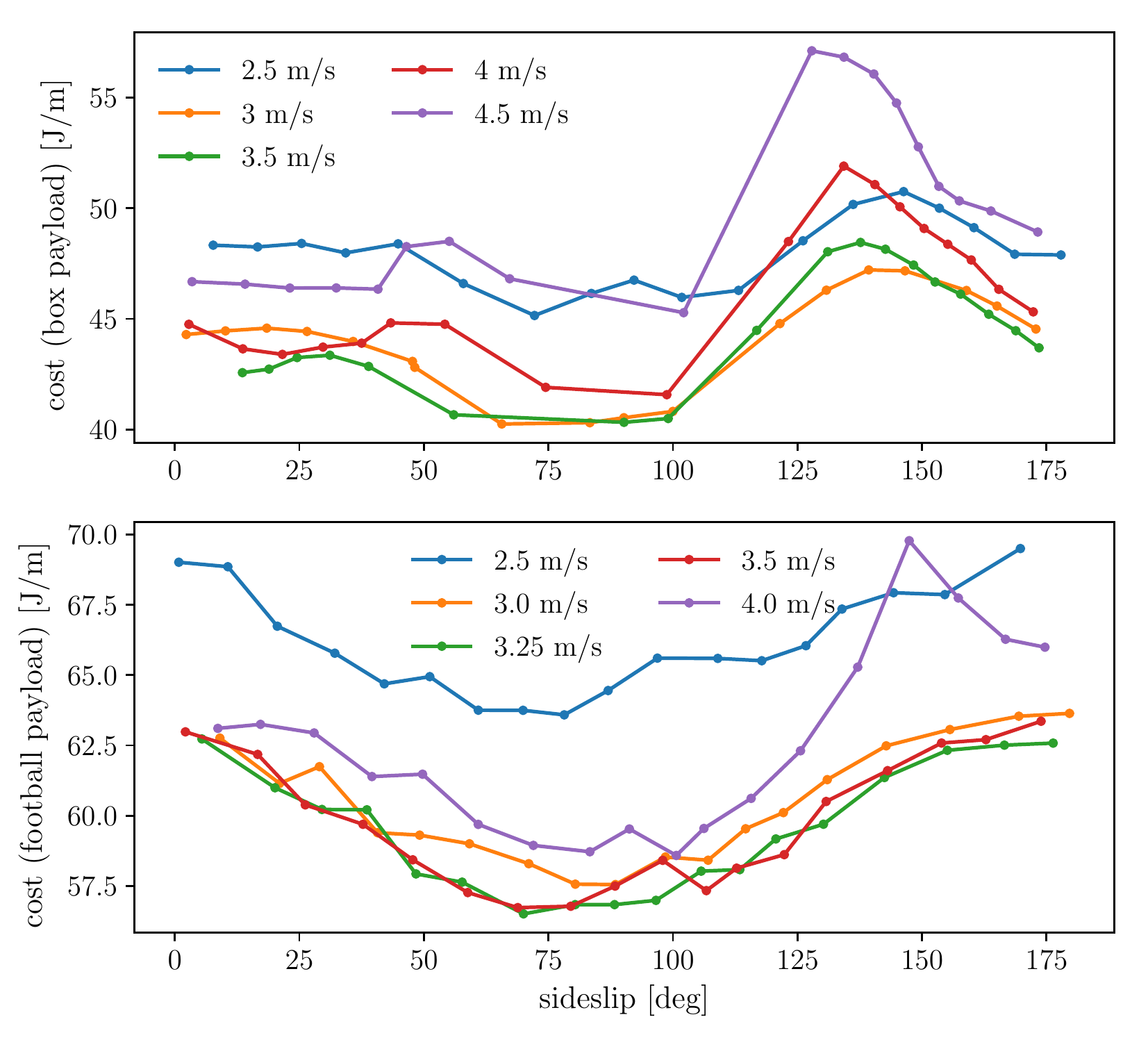}
	\caption{Ground truth values of the cost function with the cardboard box payload (the first row) and with the football payload (the second row). Each data point in this figure is the average value of 15 seconds' experimental data with a frequency of 100 Hz. The cost function reaches the minimum value when the speed is 3.0 -- 3.5 m/s and the sideslip is 65 -- 100$^\circ$ for the cardboard box payload, and  when the speed is about 3.25 m/s and the sideslip is 70$^\circ$ -- 95$^\circ$ for the football payload.}
	\label{fig:groundTruth}
\end{figure}

In the experiments, the quadcopter was commanded to fly along a circular path with constant height and a radius of 1.7 meters due to the space constraint. The cost function used is derived in (\ref{eq:RangeCostFunctionFinal}), where the power measurement $p$ was provided by the onboard voltage and current sensor and the speed $v$ and was provided by the state estimator based on the motion capture system.
To verify that the vehicle converges to the optimal speed and sideslip, we experimentally evaluated the cost function and provided the ground truth values in \figref{fig:groundTruth}.

\begin{figure}
	\centering
	\includegraphics[width=\linewidth]{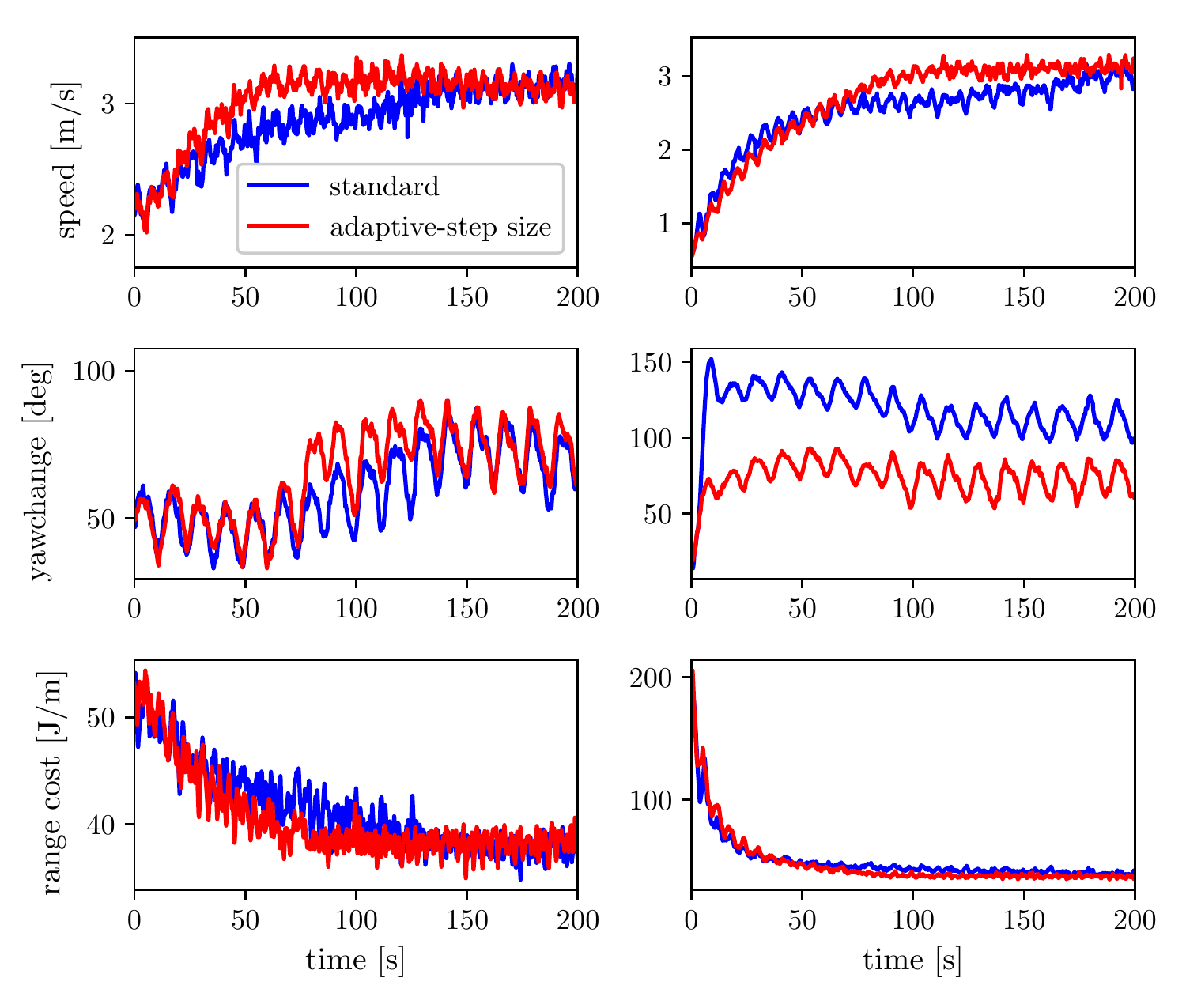}
	\caption{Comparison between the proposed method (in red) and the standard method (in blue) when using the cardboard box payload. Initial condition was 2.2 m/s and 50$^\circ$ for the first test (the first column) and was 0.5 m/s and 20$^\circ$ for the second test (the second column).}
	\label{fig:rangeCompareBox}
\end{figure}

When the cardboard box was used as payload, the performance of the proposed and standard extremum seeking controller is compared in \figref{fig:rangeCompareBox} in two tests with different initial conditions. 
In both tests, the proposed method converged in about 100 s, while the standard method took longer to converge: about 150 s in the first test and about 200 s in the second test. 
Both methods converged to speed and sideslip close to the optimal values shown in the first row of \figref{fig:groundTruth}.

\begin{figure}
	\centering
	\includegraphics[width=\linewidth]{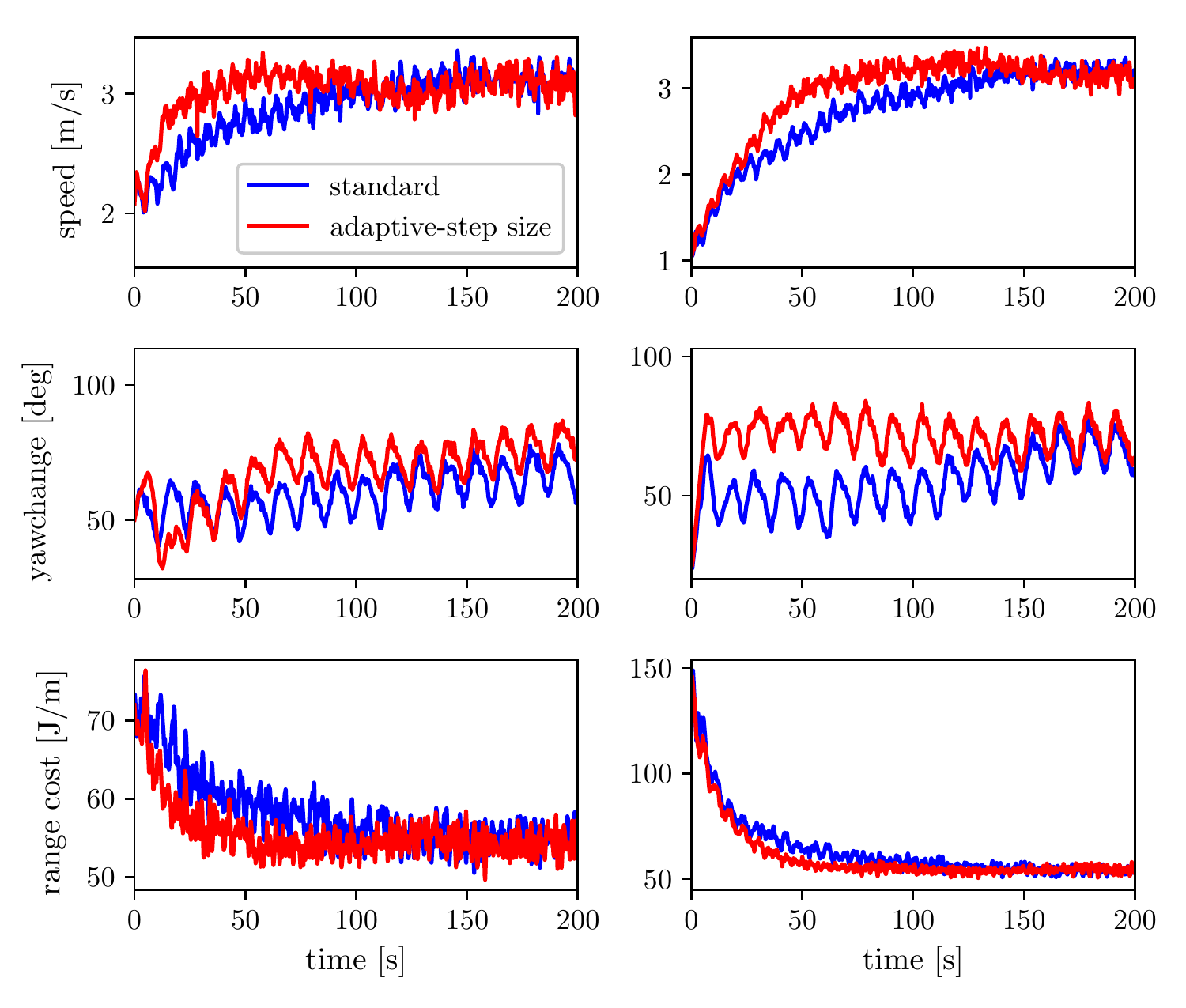}
	\caption{Comparison between the proposed method (in red) and the standard method (in blue) when using the football payload. 
	Initial condition was 2.1 m/s and 50$^\circ$ for the first test (the first column) and was 1.0 m/s and 25$^\circ$ for the second test (the second column). }
	\label{fig:rangeCompareFootball}
\end{figure}

When the football was used as payload, the performance of the two methods is compared in \figref{fig:rangeCompareFootball}.  In the first test, the proposed method converged in about 75 s and the standard method converged in about 125 s; in the second test, the proposed method converged in about 100 s and the standard method converge in about 150 s.
Both methods converged to speed and sideslip close to the optimal values shown in the second row of \figref{fig:groundTruth}.

The experiments have shown that the proposed method can find the optimal range speed and sideslip despite different payloads and initial conditions, and converges more than $30 \%$ faster compared to the standard method.

%% file: Sec5-Conclusion.tex
An online, adaptive method for finding the speed and sideslip that maximize the flight range of multicopters is proposed in this work.
Not dependent on any power consumption model of the vehicle, it can adapt to disturbances such as different payloads and is easy to deploy. 
The proposed method can mitigate the common problem of limited flight range of multicopters and thus improve their autonomy.
Based on a novel multivariable extremum seeking controller with adaptive step size, it can achieve faster convergence compared to the standard extremum seeking controller with fixed step size.
In addition, compared to our previous work of searching for the optimal speed only \cite{previousWork}, this work takes finding the optimal sideslip into account and is able to further improve the flight range.
Through indoor experiments, we show that this method is able to find the optimal speed and sideslip correctly under different payloads and initial conditions, and it converges more than $30\%$ faster compared to the standard method.

In future work, we plan to do outdoors experiments to further validate the effectiveness of this method, where accurate state estimation from the motion caption system is not available. In addition, the vehicle can fly a circular path with a much larger radius outdoors, which reduces the effect of centripetal force on power consumption.

%% file: main.bbl
\begin{thebibliography}{10}
\providecommand{\url}[1]{#1}
\csname url@samestyle\endcsname
\providecommand{\newblock}{\relax}
\providecommand{\bibinfo}[2]{#2}
\providecommand{\BIBentrySTDinterwordspacing}{\spaceskip=0pt\relax}
\providecommand{\BIBentryALTinterwordstretchfactor}{4}
\providecommand{\BIBentryALTinterwordspacing}{\spaceskip=\fontdimen2\font plus
\BIBentryALTinterwordstretchfactor\fontdimen3\font minus
  \fontdimen4\font\relax}
\providecommand{\BIBforeignlanguage}[2]{{%
\expandafter\ifx\csname l@#1\endcsname\relax
\typeout{** WARNING: IEEEtran.bst: No hyphenation pattern has been}%
\typeout{** loaded for the language `#1'. Using the pattern for}%
\typeout{** the default language instead.}%
\else
\language=\csname l@#1\endcsname
\fi
#2}}
\providecommand{\BIBdecl}{\relax}
\BIBdecl

\bibitem{AerialPhotograpy}
\BIBentryALTinterwordspacing
Y.~Ham, K.~K. Han, J.~J. Lin, and M.~Golparvar-Fard, ``Visual monitoring of
  civil infrastructure systems via camera-equipped unmanned aerial vehicles
  (uavs): a review of related works,'' \emph{Visualization in Engineering},
  vol.~4, no.~1, p.~1, Jan 2016. [Online]. Available:
  \url{https://doi.org/10.1186/s40327-015-0029-z}
\BIBentrySTDinterwordspacing

\bibitem{tagliabue2017collaborative}
A.~Tagliabue, M.~Kamel, S.~Verling, R.~Siegwart, and J.~Nieto, ``Collaborative
  transportation using mavs via passive force control,'' in \emph{Robotics and
  Automation (ICRA), 2017 IEEE International Conference on}.\hskip 1em plus
  0.5em minus 0.4em\relax IEEE, 2017, pp. 5766--5773.

\bibitem{bahnemann2017decentralized}
R.~B{\"a}hnemann, D.~Schindler, M.~Kamel, R.~Siegwart, and J.~Nieto, ``A
  decentralized multi-agent unmanned aerial system to search, pick up, and
  relocate objects,'' \emph{arXiv preprint arXiv:1707.03734}, 2017.

\bibitem{Inspection}
\BIBentryALTinterwordspacing
A.~Bircher, M.~Kamel, K.~Alexis, M.~Burri, P.~Oettershagen, S.~Omari,
  T.~Mantel, and R.~Siegwart, ``Three-dimensional coverage path planning via
  viewpoint resampling and tour optimization for aerial robots,''
  \emph{Autonomous Robots}, vol.~40, no.~6, pp. 1059--1078, Aug 2016. [Online].
  Available: \url{https://doi.org/10.1007/s10514-015-9517-1}
\BIBentrySTDinterwordspacing

\bibitem{karydis2017energetics}
K.~Karydis and V.~Kumar, ``Energetics in robotic flight at small scales,''
  \emph{Interface focus}, vol.~7, no.~1, p. 20160088, 2017.

\bibitem{TriangularQuad}
S.~Driessens and P.~Pounds, ``The triangular quadrotor: A more efficient
  quadrotor configuration,'' \emph{IEEE Transactions on Robotics}, vol.~31, pp.
  1--10, 10 2015.

\bibitem{kalantari2013design}
A.~Kalantari and M.~Spenko, ``Design and experimental validation of hytaq, a
  hybrid terrestrial and aerial quadrotor,'' in \emph{Robotics and Automation
  (ICRA), 2013 IEEE International Conference on}.\hskip 1em plus 0.5em minus
  0.4em\relax IEEE, 2013, pp. 4445--4450.

\bibitem{switchingBattery}
K.~P. Jain and M.~W. Mueller, ``Flying batteries: In-flight battery switching
  to increase multirotor flight time,'' \emph{arXiv preprint arXiv:1909.10091},
  2019.

\bibitem{modelBased1}
N.~{Bezzo}, K.~{Mohta}, C.~{Nowzari}, I.~{Lee}, V.~{Kumar}, and G.~{Pappas},
  ``Online planning for energy-efficient and disturbance-aware uav
  operations,'' in \emph{2016 IEEE/RSJ International Conference on Intelligent
  Robots and Systems (IROS)}, Oct 2016, pp. 5027--5033.

\bibitem{modelBased2}
F.~{Morbidi}, R.~{Cano}, and D.~{Lara}, ``Minimum-energy path generation for a
  quadrotor uav,'' in \emph{2016 IEEE International Conference on Robotics and
  Automation (ICRA)}, May 2016, pp. 1492--1498.

\bibitem{ampatis2014parametric}
C.~Ampatis and E.~Papadopoulos, ``Parametric design and optimization of
  multi-rotor aerial vehicles,'' in \emph{Applications of Mathematics and
  Informatics in Science and Engineering}.\hskip 1em plus 0.5em minus
  0.4em\relax Springer, 2014, pp. 1--25.

\bibitem{uavEnergyModeling}
N.~{Michel}, A.~K. {Sinha}, Z.~{Kong}, and X.~{Lin}, ``Multiphysical modeling
  of energy dynamics for multirotor unmanned aerial vehicles,'' in \emph{2019
  International Conference on Unmanned Aircraft Systems (ICUAS)}, June 2019,
  pp. 738--747.

\bibitem{leishman2006principles}
G.~J. Leishman, \emph{Principles of helicopter aerodynamics}.\hskip 1em plus
  0.5em minus 0.4em\relax Cambridge university press, 2006.

\bibitem{kreciglowa2017energy}
N.~Kreciglowa, K.~Karydis, and V.~Kumar, ``Energy efficiency of trajectory
  generation methods for stop-and-go aerial robot navigation,'' in
  \emph{Unmanned Aircraft Systems (ICUAS), 2017 International Conference
  on}.\hskip 1em plus 0.5em minus 0.4em\relax IEEE, 2017, pp. 656--662.

\bibitem{multiESC1}
K.~B. {Ariyur} and M.~{Krstic}, ``Analysis and design of multivariable extremum
  seeking,'' in \emph{Proceedings of the 2002 American Control Conference (IEEE
  Cat. No.CH37301)}, vol.~4, May 2002, pp. 2903--2908 vol.4.

\bibitem{multiESC2}
M.~A. {Rotea}, ``Analysis of multivariable extremum seeking algorithms,'' in
  \emph{Proceedings of the 2000 American Control Conference. ACC (IEEE Cat.
  No.00CH36334)}, vol.~1, no.~6, June 2000, pp. 433--437 vol.1.

\bibitem{Adam}
D.~P. Kingma and J.~Ba, ``Adam: A method for stochastic optimization,''
  \emph{CoRR}, vol. abs/1412.6980, 2015.

\bibitem{previousWork}
A.~{Tagliabue}, X.~{Wu}, and M.~W. {Mueller}, ``Model-free online motion
  adaptation for optimal range and endurance of multicopters,'' in \emph{2019
  International Conference on Robotics and Automation (ICRA)}, May 2019, pp.
  5650--5656.

\bibitem{schulz2015high}
M.~Schulz, F.~Augugliaro, R.~Ritz, and R.~D'Andrea, ``High-speed, steady flight
  with a quadrocopter in a confined environment using a tether,'' in
  \emph{Intelligent Robots and Systems (IROS), 2015 IEEE/RSJ International
  Conference on}.\hskip 1em plus 0.5em minus 0.4em\relax IEEE, 2015, pp.
  1279--1284.

\bibitem{ware2016analysis}
J.~Ware and N.~Roy, ``An analysis of wind field estimation and exploitation for
  quadrotor flight in the urban canopy layer,'' in \emph{Robotics and
  Automation (ICRA), 2016 IEEE International Conference on}.\hskip 1em plus
  0.5em minus 0.4em\relax IEEE, 2016, pp. 1507--1514.

\end{thebibliography}
